\documentclass[10pt, a4paper]{article}

\usepackage[final]{lrec2026} 

\usepackage{xcolor}
\usepackage[normalem]{ulem}
\usepackage{adjustbox}
\usepackage{stfloats}
\usepackage{booktabs}
\usepackage{tabularx,booktabs,array}
\usepackage{xcolor}
\usepackage{multirow}

\usepackage{enumitem}
\usepackage{ragged2e}

\newcommand{\datasetName}{RadTimeline}

\newcolumntype{C}{>{\centering\arraybackslash}X}
\newcommand{\question}[1]{\textcolor{blue}{#1}}
\renewcommand{\question}[1]{}

\newcommand{\unsure}[1]{\textcolor{red}{#1}}
\renewcommand{\unsure}[1]{{#1}}

\newcommand{\embPrefix}{Represent the radiological findings and be aware of the type and location of the findings}
\title{RadTimeline: Timeline Summarization for Longitudinal Radiological Lung Findings }

\name{Sitong Zhou, Meliha Yetisgen, Mari Ostendorf} 

\address{University of Washington \\
         \{sitongz, melihay, ostendor\}@uw.edu\\}

\abstract{
Tracking findings in longitudinal radiology reports is crucial for accurately identifying disease progression, and the time-consuming process would benefit from automatic summarization. This work introduces a structured summarization task, where we frame longitudinal report summarization as a timeline generation task, with dated findings organized in columns and temporally related findings grouped in rows. This structured summarization format enables straightforward comparison of findings across time and facilitates fact-checking against the associated reports. 
The timeline is generated using a 3-step LLM process of extracting findings, generating group names, and using the names to group the findings.
To evaluate such systems, we create RadTimeline, a timeline dataset focused on tracking lung-related radiologic findings in chest-related imaging reports.
Experiments on RadTimeline  show tradeoffs of different-sized LLMs and prompting strategies. Our results highlight that group name generation as an intermediate step is critical for effective finding grouping. The best configuration has some irrelevant findings but very good recall, and grouping performance is comparable to human annotators.
\\ \newline \Keywords{stuctured summarization, longitudinal EHRs, LLMs}

 }

\begin{document}

\maketitleabstract

\section{Introduction}

Tracking changes in status of findings in longitudinal radiology reports is critical for understanding disease progression, treatment response, and supporting disease risk assessment.  Reading longitudinal reports can capture subtle changes that may be missed in isolated reports. Individual reports often document changes in status with short phrases (e.g., "improved", "persists") \citep{radgraph2, namu-radie}, but the longitudinal sequence of reports can provide details that make this information more useful. 
However, reading longitudinal records and pulling together related findings can be time consuming and contributes to cognitive overload. Artificial intelligence (AI) can support this work, both in the identification and presentation of relevant findings. 

While AI tools for summarization are effective for many tasks, most work on multiple documents has involved unstructured summaries, which complicates fact-checking of statements about temporal changes \citep{chien2024aiAssistSummiarzationRadiologicReport}.
For high-stakes clinical scenarios, the ability to confirm source details behind an AI statement is critical. \citet{verma2025verifiableSummarization} propose a verifiable structured summarization format in which each fact is linked to its source document and organized into human-curated topics. 
A pre-defined set of topics is useful for large document collections, but this is not practical for summarizing radiological findings for specific patients, where it may be useful to track multiple nodules observed in the same region of the lung, for example. 

\begin{figure*}[htbp]           
  \centering                   


  \includegraphics[
      width=1.0\linewidth,     
      trim={0cm 5cm 1cm 4.5cm},
      clip                     
    ]{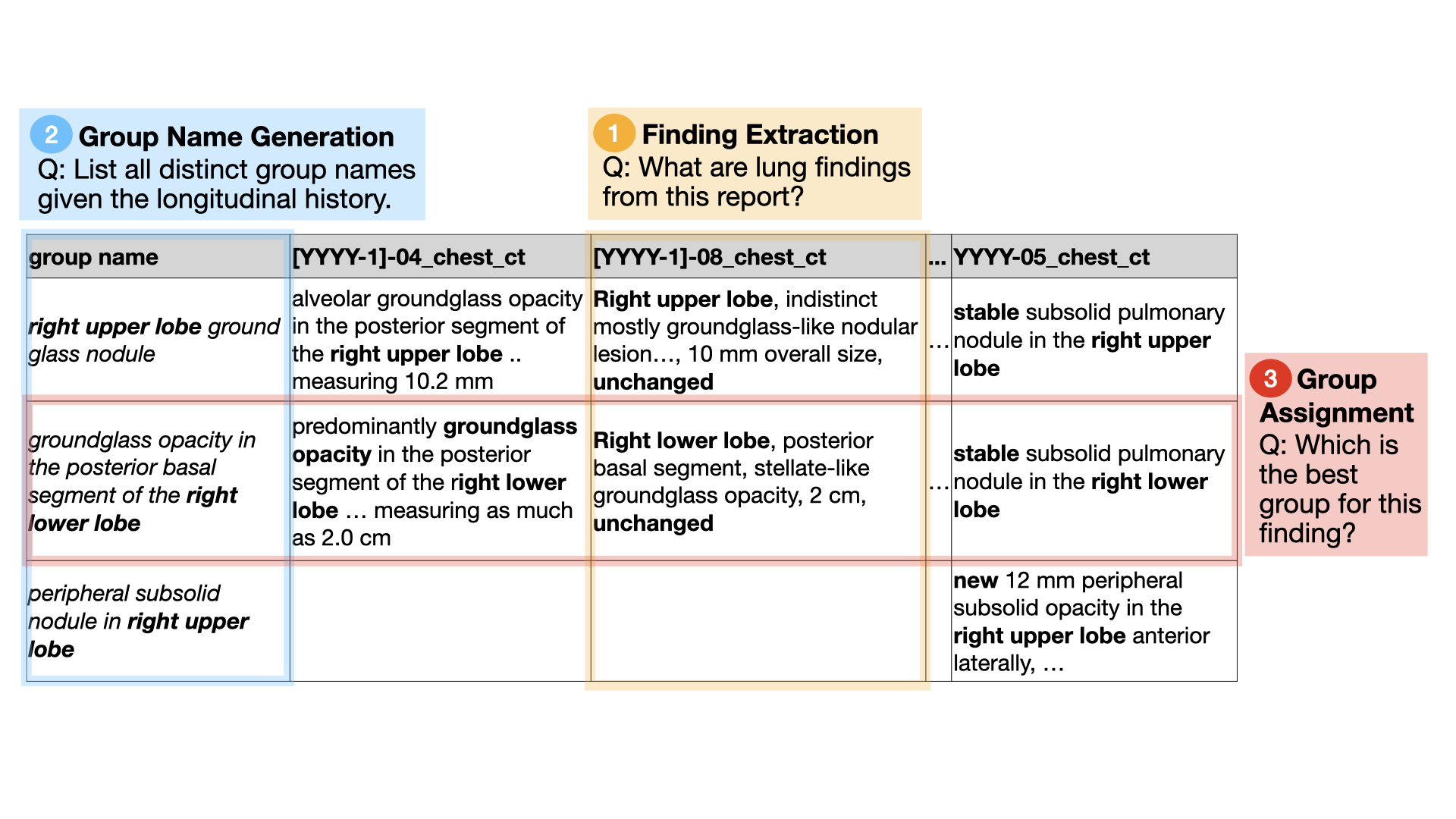}
  \caption{The timeline task and a three-step LLM approach. Each column corresponds to a time-stamped radiology exam (e.g., YYYY-05\_chest-ct). Each cell in a column is a piece of lung finding fact (e.g., "stable subsolid pulmonary nodule in the right upper lobe") from that report including clinically important details (e.g., "stable", "subsolid").  Each row groups temporally related findings, with a row header describing the distinguishing characteristics of that group (e.g., "right upper lobe ground glass nodule").}
  \label{fig:timeline}
\end{figure*}

In this paper, we introduce a new {\bf two-dimensional timeline format for structured summarization} that makes it easy to see temporal changes. 
As shown in Figure~\ref{fig:timeline} for lung-related findings, each column corresponds to a time-stamped radiology exam (as indicated in the header), such as a chest CT taken in May of a given year. Each cell in a column corresponds to a lung finding fact from that report, including details relevant to lung cancer risk assessment (e.g., size, location, status, etc.), such as ''stable subsolid pulmonary nodule in the right upper lobe.'' Each row represents a group that tracks the evolution of a particular finding over time, with a row header describing the distinguishing characteristics of that group.

To support research with this framework, we create {\bf \datasetName, an evaluation dataset} with lung-related radiological findings within longitudinal chest-related imaging reports where each patient is associated with a hand-corrected timeline. In addition, metrics are proposed for assessing automatically generated timelines in terms of finding factuality, group name quality, and finding grouping performance. 

We propose a {\bf 3-step LLM-based approach to timeline generation} that: (i) extracts lung-related finding facts from individual reports (the cells in a column), (ii) generates interpretable group names (row headers) from longitudinal reports, and (iii) associates each finding with its corresponding group (row).

In experiments with \datasetName, we show that this 3-step timeline generation approach detects most lung findings correctly, but further work is needed to address   irrelevant findings.
We observe that group name quality is critical for effective grouping, particularly for weaker LLMs and embedding-based grouping strategies.
The best prompting configuration reaches near-human performance in grouping findings.

\section{Literature Review }
\subsection{Radiology report summarization
}

LLMs have been applied to summarize the findings sections of individual radiology reports into impressions, effectively capturing critical findings.   \citet{van-veen-etal-2023-radadapt} apply LoRA fine-tuning to CLIN-T5-LARGE, and human evaluations show that the model generally captures critical findings but occasionally hallucinates details or infers prior history not in the report.   \citet{adaptedLLMOutperformHuman}  demonstrate GPT-4 with in-context examples can generate more complete summaries with fewer errors than expert-written impressions, according to human judgment. \citet{radCouncilImpressionMultiAgent} add in-context examples for Llama 3 70B, and observe improved automatic metrics such as  ROUGE-L, but more factual errors according to human judgment.  Revising these summaries with an LLM can correct presence errors but has minimal effect on errors related to progression status. 


  All the works mentioned above   report automatic summarization metrics (BLEU, ROUGE-L, BERTScore) together with human evaluation, such as  Likert-scale ratings  on completeness and factuality \citep{van-veen-etal-2023-radadapt}.
  However, those automatic summarization metrics can correlate poorly with human judgments \citep{van-veen-etal-2023-radadapt}. For LLM text generation evaluation,  fine-grained factuality evaluation systems \citep{min-etal-2023-factscore, yang2024fizz} are proposed to measure on atomic fact level. They decompose generated text into atomic facts and score each fact individually. 

For longitudinal report summarization, \citet{chien2024aiAssistSummiarzationRadiologicReport} use LLMs to generate summaries directly from multiple reports. However, these unstructured summaries are difficult to verify, may omit findings and key details, and do not necessarily describe temporal changes for each finding. Structured summarization has been developed for heart failure \citep{verma2025verifiableSummarization}, linking each fact in the summary to its source document and organizing facts into human-curated, disease-specific topics. However, fixed topic structures are not well suited for radiological findings, as groups of temporally related findings vary across patients.

Different from prior work, we develop a longitudinal report summarization approach to convert a patient’s longitudinal chest imaging reports (CT, X-ray) into a timeline table, leveraging off-the-shelf LLMs and embedder models without using supervised labels. We first extract atomized lung-finding facts from the raw reports, then group temporally related facts to display the finding trajectories. 
Like some earlier work, we use ROUGE-L to evaluate findings, but we also provide scores for overgeneration of findings, quality of group names, and grouping of findings. Lastly, we compare automatic and human scoring for one LLM configuration.





\subsection{Cross-document Coreference Resolution and Clustering}

 Our tasks share similarity with Cross-document coreference (CDCR) and clustering tasks, and their related methods that inspire our approach.

\textbf{CDCR} entails detection of coreferring event mentions from multiple documents. \unsure{Our lung radiological findings can be considered as events, but we group temporally related finding facts, instead of coreferring mentions.}  CDCR is often evaluated with same-document conference metrics, MUC, B³, CEAF, their average CoNLL F1 \citep{pradhan-etal-2012-conll}. 
\citet{barhom-etal-2019-revisitingCDECR}  develop supervised neural models that model mention pairs.
 \citet{caciularu-etal-2021-cdlm-cross}  further improve the supervised results by leveraging cross-document contexts using Longformer-based models.

LLMs have been used directly for CDCR without supervised learning and require prompt engineering.  
\citet{zhao-etal-2023-CDECRInstructHumanInstructGPT} use zero-shot GPT-4 to classify pairs of decontextualized sentences of event triggers that contain document-level information, outperforming untrained crowd workers from MTurk. 
\citet{min-etal-2024-synergeticEventUnderstandingCollaborativeCDECLLM}  use GPT-4 to assign  cluster indexes to mentions and output in a JSON format, and find that GPT-4 is worse than their supervised method, and using full  contexts from all documents is worse than using only sentences containing the mentions.   
\citet{sundar-etal-2024-majorEntityIdentificationCorefLLM}  ask LLMs to replace placeholders next to the mentions with  major entity tags in long narratives.  When evaluated on gold mentions, GPT-4 is slightly better than  their supervised method but not GPT3.5.

For radiology finding tracking, 
\citet{crossDocCorefLongitudinalTrackingRad} present a CDCR dataset about tracking findings and devices across each patient’s longitudinal radiology reports, and fine-tune a BERT model for pairwise mention coreference achieving low-to-moderate  CoNLL F1  scores.  
\citet{Mathai2025PrivacyLLMFindingMatch} use zero-shot LLMs to match a sentence in a follow up report to the single best matching sentence in its prior report. 

Our group assignment step is similar to the radiology CDCR task presented by  \citet{crossDocCorefLongitudinalTrackingRad}, 
but differs in that we do not group elements via pairwise associations. Instead, we create groups by mapping elements to a group label using LLMs, either via a question-answering prompt or using a prompt that inserts group tags for sequence of findings, inspired by \citet{sundar-etal-2024-majorEntityIdentificationCorefLLM}. 
We also investigate the impact of  full longitudinal contexts, given the mixed results in previous studies \citep{caciularu-etal-2021-cdlm-cross, min-etal-2024-synergeticEventUnderstandingCollaborativeCDECLLM}.

Our task is also similar to \textbf{Clustering} in terms of grouping similar items, where  items are findings and similarity is in terms of shared characteristics among temporally related findings.
 Off-the-shelf  embedders can provide similarity measures for clustering items.
\citet{muennighoff-etal-2023-mteb} present a MTEB benchmark to evaluate text embedders on varying tasks and observe that no single embedder dominates all tasks. 
  E5-mistral-7b \citep{wang-etal-2024-improving-text-e5-mistral} is a finetuned Mistral-7B bi-encoder trained on diverse synthetic data and among the state-of-art embedder models on MTEB. They support prepending textual instructions to queries for task adaptation. 

LLMs can guide clustering by infusing task-specific preferences.
\citet{Huang2024TextClusterClassificiationLLM} use  a two–stage LLM pipeline built on GPT-3.5. They first ask LLM to generate human-interpretable group names, followed by a de-duplication step, then classify each item into a group. They find the performance gap between using generated groups and using oracle group names is small.  
This is very relevant to our LLM approaches for the group name generation and group assignment steps. Our work differs in its focus on long findings for a single patient vs. corpus-level topics.  We also collect group name reference data to evaluate the intermediate name generation step.

\section{Gold Timeline Dataset Curation }
\label{sec:data}

We sample 10 patients from a patient cohort from an existing case-control dataset created for a lung cancer surveillance project in  UW Medicine,\footnote{A subset of this cohort was previously used by \citet{zeng_traj-coa_2025} for lung cancer prediction. }
where temporal changes of lung-related  findings are clinically important risk factors. 
We ensure that each of the sampled patients having at least four chest-related radiology exams, including chest CT, chest X-ray, and abdomen CT, within a five-year window. \unsure{According to the state cancer registry, 6 patients have future lung cancer diagnosis records within 3 years, and 4 patients have no diagnosis records of any type of cancer.} The raw EHRs contain 65 chest-related exam reports from all patients. \unsure{The number of reports per patient ranges from 4 to 12, and longitudinal report lengths measured in tokens  range from 646  to 4625, with an average at 2055.}

We create a timeline for each patient based on their longitudinal radiology reports, as illustrated in Figure 1. Each column represents the findings (\textbf{gold findings}) extracted from an individual radiology report, while each row captures the temporal evolution of a specific finding mentioned across multiple reports (\textbf{gold group assignments}). Each row is further assigned a \textbf{gold group name}, summarizing the key characteristics of the findings in this row. Since our focus is on tracking lung-related conditions, findings pertaining to other organs (e.g., the aorta) or descriptions of medical devices are considered irrelevant and are omitted from the timelines.
We work with two medical students and one medical doctor as annotators during the timeline creation process. We first automatically generate timelines using LLMs (see  Section~\ref{sec:method} for details) and hand-correct LLM generated findings and group assignments based on the feedback from two medical student annotators.
\unsure{Finally, a third annotator reviews the gold finding groups, and hand-corrects row names of each version of gold group assignments to produce the gold group names.}

Annotation guidelines are included in Appendix~\ref{section:annotation_guideline}.

\subsection{Gold Findings (Columns) }
\label{sec:data:gold-finding}

\textbf{Factuality evaluation:} We ask two medical student annotators to rate the factuality of \unsure{each generated finding} against its associated report, and to add any missing findings that the model fails to include. In addition, they provide  notes explaining the identified errors, which are later used by an author to manually correct the generated findings as the gold-standard reference.

To reduce annotation efforts and focus on abnormalities, we manually drop absent findings, such as ''no new nodules,''  and normal findings, such as ``clear lungs.''   The original automatically generated timelines contain 221 findings from 61 reports. After removing absent findings,  we obtain 155 findings from 55 reports for the annotation tasks in factuality evaluation. 

We define four categories for factuality rating: "irrelevant,'' "wrong,'' "partially correct," and "correct." "Irrelevant" refers to findings that are not related to lung radiological observations, typically describing findings from other organs or medical devices. "Wrong" refers to findings that contain errors regarding  presence or temporal change trends. "Partial" is used for findings that contain inaccurate or incomplete details.
\unsure{For example, in "stable subsolid pulmonary nodule in the right upper lobe", annotators will consider it as "partial" if  "stable" or "subsolid" are missing.} 
Findings that include correct and complete clinically important details are labeled as "Correct." Table \ref{tab:confusion} shows the confusion matrix between two annotators' ratings.    The two annotations reach a \unsure{high agreement} at 86\%.
No finding is considered wrong by both of the annotators, and the major mistake is including irrelevant findings (14\%).
In addition, \unsure{one annotator} identifies 3 missing findings from a single report.
No other missing findings are reported.

\renewcommand{\arraystretch}{1.1}

\begin{table}[h]
\centering{
\setlength{\tabcolsep}{1pt} 
\begin{tabularx}{0.48\textwidth}{lXXXX }
\toprule
 & Wrong & Partial & Correct & Irrelevant \\
\midrule
Wrong        & 0 & 0  & 2   & 0  \\
Partial      & 0 & 12 & 3   & 0  \\
Correct      & 2 & 9  & 106 & 2  \\
Irrelevant \hspace{5pt}   & 1 & 0  & 2   & 16 \\
\bottomrule
\end{tabularx}}
\caption{Confusion‑matrix of factuality annotation between two annotators. Annotators agree on factuality ratings for 86\% of 155 generated findings.
}
\label{tab:confusion}
\end{table}

\textbf{Gold Finding Curation:}  We refine the LLM-generated findings as follows. Irrelevant or wrong findings are removed, correct findings are kept unchanged, and partially correct findings are either revised by adding missing details, or merged with another finding with the complementary details when they are both described in proximity within the same paragraph.  Additionally, we include the three missing findings identified during review. If annotators rate a finding differently, the first author resolves the disagreements based on annotators' notes to decide the action.  One exception is made for one finding labeled as correct by both annotators but later found to be irrelevant during post-analysis and removed from the gold references. 
For four partially correct findings in which incomplete details result from complementary information spread across different parts of a radiology report, we create a new finding and add a same-report link between the new finding and the original partial finding during the subsequent group assignment curation process.
This is consistent with the automatically generated findings, where finding details from different parts of reports are not always merged.

After resolving  disagreements, we find that 15\% of the 155 findings used in the factuality human evaluation can be dropped due to being wrong or irrelevant and 12\% could be further merged with another finding in the same report.
After removing and adding findings based on ratings and notes from annotators, there are 136 gold findings, or 115-116 merged findings after merging same-report linked findings grouped separately by the two annotators.

\subsection{Gold Group Assignments (Rows) }
To collect group annotations for temporally related findings, annotators assign a group ID to each finding. We provide them with automatically generated group names linked to the corresponding group IDs, displayed as the row headers in the generated timelines. Annotators may add new group names and IDs if the existing list is incomplete or not representative of a group. When two group IDs convey the same meaning, annotators are instructed to select one ID and apply it consistently across all relevant findings. Revising the group names is not required at this stage. Annotators have access to the full longitudinal radiology reports while performing group ID assignment. 
For the seven added findings (3 missing, 4 due to missing details), group IDs have not been pre-assigned. To assign IDs to these findings, one annotator labels the seven findings in their own group assignment version, and the first author propagates these labels to another annotator’s version to maintain consistency.

We collect two versions of group assignments to 
capture individual annotator preferences. For example, for "low lung volumes bilaterally", and "bibasilar opacities, likely atelectasis", one annotator groups them and the other separates them into different groups. The CoNLL F1 between two gold group assignments is 82, reflecting good inter-annotator agreement. 
Group annotation statistics are in Table \ref{tab:data_stats}.

\begin{table}[]

\begin{tabularx}{0.5\textwidth}{lXX}
\toprule
Statistic        & Ann-1 & Ann-2  \\
\midrule
\# \textbf{Groups}                    & 65                                 & 62                                 \\
\hspace{1.5em} Avg. / Max   size & 2.1 / 8 & 2.2 / 7 \\
\hspace{1.5em} \% Non-singleton       & 52                              & 50                             \\
 \# \textbf{Findings} (Total \# 136  )       &   &          \\

\hspace{1.5em}  \%   Linked (Any / Cross)   & 77 / 70                              & 77 / 71   \\
  \# \textbf{Findings} ( Merg.)       &       116                        &   115                  \\
\bottomrule
\end{tabularx}
\caption{Group annotation statistics from two annotators (Ann-1 and  Ann-2) on findings in \datasetName.  The group size is based on the 136 findings. "Findings (Merg.)" are findings that are merged with same-report links \unsure{for finding extraction evaluation}. The \% Linked refer to the percentage of findings that have links to any other finding  (Any) vs. only cross-note findings  (Cross) among the total 136 findings.
}
\label{tab:data_stats}
\end{table}



\subsection{Gold Group Names (Row Headers) }
\unsure{The third annotator who is a medical doctor reviews both versions of gold finding groups. The annotator is asked to manually correct row names of each version of gold group assignments to produce distinguishable group names, and each name should cover the shared information within the group. }

\section{Methods}
\label{sec:method}
We use a three-step approach to generate timelines. First, lung-related findings are identified for each report. Second, given the longitudinal report contexts, a list of group names is generated to represent distinct findings.
Lastly,  each finding is assigned to a group, optionally selected from  already generated group names. This section describes the different configurations explored for each step, including the version used in automatic timeline generation for annotation. We evaluate two LLMs, Llama 3.1 8B instruct \citep{grattafiori2024llama}, referred to as Llama 3.1, and GPT-4o \citep{hurst2024gpt4o}.\footnote{The LLM prompts for each step are provided in Appendix~\ref{section:prompt}.} 
We also evaluated GPT-OSS in preliminary experiments but do not include it as we observed no performance gains over GPT-4o.

\question{(Question: terminology, system or approach, I use them interchangeably so far. "system" has been used in multiple places in  abstract and intro, do we change them all)}

\subsection{Step 1: Finding Extraction}

We first extract lung-related findings from each report independently using  an LLM. \unsure{The prompt includes the full texts from a single report, detailed instructions about generating the lung findings, a short output example, and finally a request for the final answer.} The instruction asks for details that may be relevant to risk evaluation, such as size and change. Because lung finding descriptions can be lengthy, we require the output format as a Python list in order to simplify post-processing.  

\question{Question: check if the prompt description self-contained without Appendix, and not having too much prompt details to be confusing}

\unsure{When generating the timelines for annotation, we use Llama 3.1 in Step 1.}

We automatically remove findings that contain only absent events according to an information extraction model,  RadGraph \citep{delbrouck-etal-2022-improving-radgragh-reward}.\footnote{The RadGraph package is available at: \url{https://pypi.org/project/radgraph/}. Specifically, we use the model version \texttt{modern-radgraph-xl}.   } 
We use two versions of absent prediction filters. The first filters out sentences where all detected entities are absent. The second is more aggressive, dropping predictions when the core entities (e.g. lesions) are all absent even if their descriptors  (e.g. lung) are not.  For the 203 Llama-predicted findings, the first filter retains 187 findings, and the second reduces these to 165, closer to the 155 findings obtained in human labeling.

\subsection{Step 2: Group Name Generation}
\label{sec:method:group-name-generation}

Instead of directly clustering the findings, we use an intermediate step to generate interpretable group name in  natural language  first, which are used in the later group assignment step to map findings.  By comparing findings with group names, we can avoid expensive pairwise finding comparison.  Group names should be specific enough so that they can separate unrelated findings, and generic names should be avoided.

\unsure{The prompts include longitudinal report records of a patient, detailed instructions asking for temporally related lung finding groups, one good (specific) and one bad (generic) group name example, and the final request asking for a list of group names separated by commas.  }

\question{(Question: move "group name should be specific..." to the first paragraph, and add (specific) (generic) for examples in the prompts)}

\question{(Question: is this prompt description still too confusing)}

We explore two longitudinal input contexts to represent patient records; one includes the full reports, called \textbf{full context}, and the other includes only predicted findings from all reports in Step 1, called \textbf{finding-only context}. We use finding-only context to test whether filtering content into shorter lengths improves reasoning.  In both input context versions, the report contents are marked by  dates and exam titles.

\unsure{We use Llama 3.1 with full context in Step 2 to generate the timelines for annotation.}

\question{(Question: check for prompt self-contained without  Appendix))}

\subsection{Step 3: Group Assignment}
In this step, we assign each finding to a group name, so that findings with the same group name are grouped. We propose both LLM-based and more efficient embedding-based methods, each of which leverages the group names predicted in Step 2.

\subsubsection{LLM-based Approach}
We explore \unsure{ prompting variations } including  single finding (\textbf{Single}) or  multiple findings   (\textbf{Multiple})  per LLM run, zero-shot (\textbf{ZS}) or  few-shot  (\textbf{FS}), and using full longitudinal contexts (\textbf{full context}) or \textbf{no context}.

For \textbf{Single} prompting, \unsure{the LLM is prompted to assign each finding individually to the group  best describing this finding among the provided group name list, or “Other”  if no suitable group exists.}  The group name list is predicted in Step 2 for the same patient. If the prediction is 'Other', the finding's own text is used as its group name. The group names from Step 2 that are not assigned to any findings will be dropped from the final timelines.   We instruct the LLM to place the selected group name in the last line of its response. During post-processing, the first group name appearing in the last line is parsed as the prediction, even if multiple group names are present.

When the generated group names are incomplete or duplicated, findings from the same group may not share  consistent group names. To encourage consistent group assignment across all related findings, we experiment with \textbf{Multiple} prompting, which \unsure{assigns groups to all findings of a patient in a single LLM run by inserting  a tag next to each finding}, following the approach of \citet{sundar-etal-2024-majorEntityIdentificationCorefLLM}. Those tags serve as interpretable group names in natural language, and are parsed from angle brackets as the predicted group name for the corresponding finding.
We use two variations. One does not require existing  group names in the prompts (\textbf{no groups}),
and the other requires them so that  LLM can choose from existing group names or add new ones. 
This Multiple prompting is cheaper than Single prompting due to fewer runs.

 We experiment with using  all longitudinal radiology notes (\textbf{full context}) to provide complete cross-document information. This setup is similar to the human annotation setting where annotators have access to the full set of longitudinal reports.  However, it remains unclear whether LLMs can effectively utilize such long contexts, and processing with full context is computationally expensive. Therefore, we also evaluate \textbf{no context} prompts, where group assignment relies solely on the provided group names and the finding texts.

The few-shot (\textbf{FS}) prompts are motivated by formatting errors observed when  zero-shot (\textbf{ZS}) prompts \unsure{are used with Llama 3.1}. 
We manually curate two short  examples without using any examples from the evaluation dataset. Each example includes the existing group names if in use, and the output in the required format.

\unsure{The final prompts integrate the above variations in the following order:  longitudinal reports (omit for no context), detailed instructions (varies for Single or Multiple), two examples (omit for ZS), group names list (omit if no groups),  
a single or multiple findings to be grouped, 
and a final output request  emphasizing on the output format.}

\unsure{We use  Single ZS full-context with Llama 3.1  in Step 3 to generate  timelines for annotation.}

\question{(Question for Mari:  those prompt description is shorten  now. Is it still confusing?)}

\subsubsection{Embedder-based Approach}

We also explore embedding-based methods as more efficient alternatives for LLMs in the group assignment step. We first use a \unsure{non-generative} embedder model to convert each finding and each group name into embeddings, then assign each finding to the group name with the highest cosine similarity from the same patient. No new groups will be added beyond the provided group names.

 Our embedder backbone is E5-Mistral which is built for general domain and supports instruction integration for task adaptation. 
 We first describe the task using a \unsure{\textbf{general instruction}}, "Given a radiology finding, find the group that it belongs to", which is prepended to each finding. \unsure{We further infuse task-specific preference using a \textbf{task-specific instruction}}.  We \unsure{hypothesize} that finding types (e.g., "opacities", "nodules") and anatomy locations (e.g., "right upper lobe") are critical information that distinguishes lung radiological groups. Therefore, we add a prefix,  "\embPrefix", to both findings and group names.

\section{Experiment and Results}
We evaluate \unsure{Llama 3.1} and GPT-4o for prompting methods in all steps. Context lengths of both are 128k tokens. We consistently choose greedy decoding for Llama 3.1  and set the temperature at 0.1, top-p at 0.9 for GPT-4o, without hyperparameter tuning. The group assignment step is evaluated on gold findings.

\subsection{Finding Extraction }
\label{sec:results:step1}

We compute ROUGE-L  scores at both the report level and finding level, in both cases by first scoring against each of the two annotators and then averaging the results.
For the report-level score,  we represent a report as the concatenated list of all findings before automatically removing absent findings, compute the ROUGE-L score for the list, and then average ROUGE-L scores across reports.  
When calculating the finding-level scores, we use the merged gold findings (findings from the same report assigned to the same group) and represent them using a concatenation of the respective finding strings.
Then, we automatically align each predicted finding to the gold (merged) finding that yields the maximum ROUGE-L among all the gold findings from the same report, after filtering out absent findings. 
The finding-level ROUGE-L score is the average over the ROUGE-L scores for all matches.  

Gold findings that have no match are considered \textbf{missed}. 
We reversely match each gold finding to the prediction that yields the maximum ROUGE-L and report the percentage of unmatched predictions as \textbf{unnecessary}. 
Under an ideal alignment, unnecessary predicted findings would include wrong, irrelevant, redundant, and unfiltered absent predictions.

\question{Question: in Introduction, "but further work is needed to address duplicative and irrelevant findings. ", does the future work include human evaluation on GPT-4o? Because I am not sure if GPT-4o produce duplicated and irrelevant findings in concerning level give no human evaluation, I think automatic scores only tells its relative level to llama. }

Table~\ref{tab:results-factuality} reports the different scores for finding extraction for two LLMs, Llama 3.1 and GPT-4o.  Report-level ROUGE-L scores for Llama 3.1 and GPT-4o are close. For finding-level scoring, we provide two configurations, one with a weaker and one with a more aggressive absent finding filter.
 For both configurations,  GPT-4o outperforms Llama 3.1 (67 versus 62). GPT-4o also has substantially lower unnecessary predictions. For examples, see Appendix~\ref{section:examples} Table~\ref{tab:step1_examples}. Both LLMs have low rates of missed findings. 
As expected, the more aggressive filtering of absent findings leads to substantial reduction in the percent of unnecessary findings. In addition, the finding-level ROUGE-L score improves, with minimal increase in missed findings.
The ideal alignment of reference to gold findings would yield roughly 3\% missing and 39\% unnecessary findings for Llama 3.1 predictions, which is close to the ROUGE-L alignment.

\renewcommand{\arraystretch}{1.1}
\begin{table}[ht]
\centering
\begin{tabularx}{0.48\textwidth}{lXXXXX}
\toprule
Model & RL (Rpt.) & RL (Fdg.) & Miss. \% & Unnec. \% \\
\midrule
Llama-3.1   & 67    & 62  & 3           & 40                                             \\
GPT-4o   & 68   & 67   & 5              & 25                                 \\   
Llama-3.1*              & -             & 70     & 3      & 32     \\
GPT-4o*                & -             & 75     & 6       & 15     \\
\bottomrule
\end{tabularx}

\caption{Finding extraction evaluation against gold findings. 
ROUGE-L (RL) scores are at the report (Rpt.) and finding (Fdg.) level. Missing (Miss.) represents unmatched-gold percentage, and unnecessary (Unnec.) represents unmatched prediction percentage.
The first two rows use the weaker absent finding filter; * indicates to the more aggressive filter.
}
\label{tab:results-factuality}
\end{table}

\subsection{Group Name Generation}
As timeline row headers, generated group names are 
useful for readability, but they
also impact the later group assignment step as intermediate inputs.

Similarly to the finding-level ROUGE-L scores in Section ~\ref{sec:results:step1}, we compute the group-name-level  ROUGE-L scores, missing and unnecessary prediction rates, as well as the percentage of gold names that have multiple predicted versions (duplicates) under automatic alignment. We compare the generated group names against two versions of gold group names separately, and report the average.

Table \ref{tab:results_group_name_metrics} reports the scores for the generated group names from  two LLMs, using either full contexts or finding-only contexts, as well as the oracle scores evaluated between the two versions of gold group names.  To calculate the oracle scores, we evaluate one gold group version against the other alternatively and take the averages. When using full contexts, GPT-4o has a higher ROUGE-L than Llama 3.1 (48 versus 38), but also a higher unmatched gold percentage (30\% versus 25\%).  The Llama 3.1 model has nearly 60\%  unmatched predictions, suggesting considerably more unnecessary group names, likely due to irrelevant  and duplicative predictions. When switching to finding-only contexts, the duplication issues in Llama 3.1 (from 32\% to 42\%) and  unnecessary predictions in GPT-4o (from 37\% to 46\%)  become more severe. In contrast, the missing percentage of GPT-4o decreases from 30\% to 13\%. These results suggest that using itemized findings as contexts produces a longer list of group names. Examples are given in Appendix~\ref{section:examples} Table~\ref{tab:step2_examples} and \ref{tab:step2_examples_full_vs_list}.


 

\begin{table}[ht]
\centering
\begin{tabularx}{0.5\textwidth}{lXXXX}
\toprule
Method & RL & Miss. \%  & Dup. \%  & Unnec. \% \\

\midrule

Llama  3.1 (Full)            & 38 & 25 & 32 & 60 \\
Llama 3.1  (Fdg.)     &     42	& 27 & 	42 &	58 \\

GPT-4o  (Full)             & 48  & 30  & 28 & 37 \\
GPT-4o  (Fdg.)              &  52 &  13  &  31 & 46 \\
Oracle    & 87 & 9 & 11 & 9 \\
\bottomrule
\end{tabularx}

\caption{Evaluation results for group name generation. The groups are generated using full or finding-only (Fdg.) contexts. } 

\label{tab:results_group_name_metrics}
\end{table}

\subsection{Group Assignment}
\label{sec:results:step3}
We evaluate group assignment methods on gold findings, in order to directly compare with their associated gold group assignments.  We consider findings that share the same group name and come from the same patient as a group, and report CoNLL F1 scores as in CDCR tasks.  Given that we have two versions of gold group assignments, we calculate  CoNLL F1 against each version and report the averages. 
\unsure{We also report oracle results, assuming that gold group names are known and used as intermediate inputs, and further average the scores when evaluating against two gold group name versions separately.}

\subsubsection{Comparing Prompting and Embedder Methods without Contexts}

\begin{table*}[ht]
\centering

\begin{tabularx}{\textwidth}{l|CCCC|CCCC|CC}
\toprule
         Group names & Single ZS Llama & Single FS Llama & Multi. ZS Llama & Multi. FS Llama & Single ZS GPT & Single FS GPT & Multi. ZS GPT & Multi. FS GPT  & Embed. Gen. & Embed. Spec. \\
\midrule
Llama-3.1 & 66                                & 74                                & 77                               & 78                               & 77                                 & 73                                 & 78                                    & 77                                    & 68                                                                          & 70                                                     \\
GPT-4o  & 75                                & 79                                & 77                               & 81                               & 82                                 & 80                                 & 84                                    & 82                                    & 65                                                                          & 69                                                     \\
Oracle       & 73                                & 78                                & 86                               & 84                               & 80                                 & 84                                 & 85                                    & 83                                    & 78                                                                          & 83                                                     \\
No groups     &            n/a                       &       n/a                             & 72                               & 68                               &             n/a                        &        n/a                             & 70                                    & 72                                    &                n/a                                        &   n/a\\

\bottomrule
\end{tabularx}
\caption{Group assignment performance of prompting and embedding-based methods with no context in CoNLL F1 scores.  We use LLM models (Llama 3.1, GPT-4o) and an embedder model (e5-mistral-7b-instruct).  "Multi." is short for Multiple prompting. "Embed. Gen." and "Embed. Spec." represent embedding-based methods using the general or task-specific instructions.
}
\label{tab:results_grouping}
\end{table*}

Table \ref{tab:results_grouping} compares prompting methods and  embedding-based methods  when using no context in Step 3. We evaluate these methods when using generated group names, gold group names (oracle), or no group names. All generated group names are obtained from Step 2 using full contexts.

 \textbf{The choice of intermediate group names is crucial, and GPT-4o approaches reach near-human performance on gold findings.} When shifting group names generated by Llama 3.1 to GPT-4o,   all prompting methods improve, except that  Multiple ZS prompt with Llama 3.1 is unchanged.  
 The Single FS prompt with Llama 3.1 improves from 74 to 79 in CoNLL F1, and with GPT-4o it improves from 73 to 80.
 GPT-4o prompting methods with GPT-4o group names are all close to the inter-annotator performance at 82 CoNLL F1, and the best one, using Multiple ZS, reaches 84.
Embedding-based methods do not show improvement with GPT-4o-generated names. 
Under oracle scenarios, prompting methods show mixed results compared to using automatically generated group names, indicating that automatic names are not necessarily worse than human curated ones. In contrast, embedding-based methods significantly improve with oracle names, and the best embedding-based method, which uses task-specific instructions, achieves near-human performance at 84 CoNLL F1, indicating their potential as efficient substitutes to LLMs when group name generation is further improved.
For the Multiple prompt version, we observe a considerable performance drop to around 70 after removing group names, suggesting the importance of including the group name generation step.

\question{Question: Do we own explanation for embedder didn't improve with GPT4o? 
I  delete this llama bias explanation: "One possible explanation is that ...may bias the evaluation toward Llama 3.1’s group names due to similar language styles." -> }

 \textbf{Domain knowledge helps Embedding-based methods}. We compare the task-specific instruction and the general instruction for embedding-based methods. It shows that adding domain knowledge to instructions that emphasize the finding types and anatomy locations consistently  boost results under all group name settings ( from 78 to 83 with oracle names). The gap to LLM-based methods is narrower but still persists. More task adaptation efforts on embedding-based methods are needed.

 \textbf{Few-shot prompting only helps Llama 3.1}. We observe that  few-shot examples help  Llama 3.1 in following a specified format. \unsure{However, for GPT-4o, few-shot examples do not seem necessary.}

\subsubsection{Impact of Full Contexts}
\begin{table}[ht]
\centering

\begin{tabularx}{0.5\textwidth}{lXXXXXX}
\toprule
 Context  & Single ZS Llama  & Single FS Llama   & Multi. ZS Llama  & Multi. FS Llama  & Single ZS  \hspace{10pt} GPT  & Multi. ZS  \hspace{10pt}  GPT   \\
  \midrule
 Ann1  &  &  &  &  &  &  \\
No Ctx.   & 67 & 74 & 79 & 78 & 81 & 82 \\
Full Ctx.  & 56 & 70 & 76 & 78 & 77 & 81 \\
\midrule
  Ann2   &  &  &  &  &  &  \\
No Ctx.   & 65 & 75 & 75 & 77 & 82 & 85 \\
Full Ctx. & 60 & 76 & 82 & 78 & 86 & 90 \\
\bottomrule
\end{tabularx}
\caption{Impact of full contexts in group assignment prompting methods. Evaluated in CoNLL-F1 against  gold group assignments from two annotators separately. Each configuration uses either Llama 3.1 for all steps or uses GPT-4o for all. All experiments use full contexts by default in group name generation. "Ctx." is short for "Context". }

\label{tab:results_llm_contexts}
\end{table}

Table \ref{tab:results_llm_contexts}  shows the impact of adding full contexts for group assignment prompts, evaluated against two versions of gold group assignments separately.
With the exception of Llama 3.1 Single ZS prompting, where full context hurts using both gold annotations, full context improves performance when measured against gold group assignments from  annotator 2 but not annotator 1.
\unsure{These results suggest annotator differences in leveraging long-contexts, which is worth future investigation. }

\begin{table}[ht]
\centering
\begin{tabularx}{0.5\textwidth}{lXXXX}
\toprule
           Group names                              & Single  ZS  & Single  FS   & Multi.  ZS  & Multi.  FS  \\
\midrule
Llama  (Full)         & 66 & 74 & 77  & 78  \\
Llama  (Fdg.) & 66 & 70  & 71 & 68 \\
\midrule

GPT (Full)           & 82 & 80 & 84 & 82  \\
GPT (Fdg.)    &  73 &  74 & 74 &  73 \\
\bottomrule
\end{tabularx}
\caption{Impact of input contexts for group name generation  on the final  group assignment performance in CoNLL-F1.  All experiments use the same LLM in all steps and no contexts in group assignment. Group names are generated using full or finding-only ("Fdg.") contexts.}
\label{tab:results_group_method_variation}
\end{table}
In Table \ref{tab:results_group_method_variation}, we further evaluate the impact of different group name generation prompts on group assignment. We observe that switching from full contexts to finding-only contexts for group name generation hurts the final group assignment performance, indicating that the additional context is useful. 

\section{Conclusion}
\question{Question: I plan to finally compile with lrec2026-example\_migrate\_later.tex instead of the current main\_lrec-coling2024-example.tex file. I compare two versions of pdfs before and didn't notice a difference, but LREC specifies the 2026 templates, so  I feel it is better to use exact the same template.  let me know if you have different opinions.}

We propose a timeline generation task for structured summarization on longitudinal radiology reports. This timeline format groups temporally related findings for straightforward comparison and facilitates fact-checking by denoting  the associated report for each finding.
 We create a timeline dataset, RadTimeline, to evaluate timeline generation. We propose a three-step LLM approach that can achieve near-human performance in grouping gold findings.

This work can be extended to a broader clinical longitudinal records to capture temporal nuances in clinical narratives, such as for symptom progression, medicine history, and treatment trajectories. 

\section{Ethics}
We receive approval from the Institutional Review Board (IRB) to access the patient data.
All code and data are stored and executed on servers that complies with the Health Insurance Portability and Accountability Act (HIPAA). We use a HIPAA-compliant version  of APIs for GPT-4o experiments.
All authors involved in  running code and accessing the database have completed  human subjects training. Our dataset will be de-identified to remove Protected Health Information (PHI) using an in-house tool \citep{deidTool}. Release of the de-identified dataset is contingent on institutional approval.

\section{Limitations}

We report performance with only Llama 3.1 8B Instruct and GPT-4o. Due to data privacy concerns, our choice of proprietary LLMs is limited to those with HIPAA-compliant APIs.

When evaluating finding extraction and group name generation, we align predictions and gold references using a maximum ROUGE-L criterion.
Using the ROUGE-based automatic alignment to score the gold findings give a rate of unnecessary predictions for Llama 3.1 in Table~\ref{tab:results-factuality} that is consistent with aggregate statistics in the human annotation of those predictions. However, a more fine-grained analysis is needed to assess quality of the alignments.

Our method for evaluating grouping of findings requires that the findings have gold group assignments, which are only available for the gold findings.
One solution would be to assign "silver" labels to the predicted findings by aligning them to gold findings, as in the finding assessment approach.
Of course, this strategy should be validated through human evaluation of the fully automatic timeline.

The evaluation dataset contains only 10 patients, due to the costs associated with clinical excerpts and the cognitive effort required to understand longitudinal reports.
This evaluation dataset gold annotations may share a similar language style to  Llama 3.1, because the dataset is hand-corrected from generated timelines produced by Llama 3.1 based on expert feedback. This could lead to bias in ROUGE-L score.

We will de-identify the longitudinal reports when releasing the dataset.  The performance may change  after de-identification, and we will rerun our best methods to report performance on the de-identified dataset upon release.

\section{Acknowledgment}
The authors thank the annotators Emily Stiles, Nianiella Dorvall, Zixuan Yu  for their contributions to this work. This work was supported in part by the National Institutes of Health (NIH) National Cancer Institute (NCI) (GrantNr. 1R01CA248422-01A1) and NIH Clinical and Translational Science Awards Program (CTSA) (GrantNr.UL1 TR002319, KL2 TR002317, and TL1 TR002318). The content is solely the responsibility of the authors and does not necessarily represent the official views of the National Institutes of Health.

\section{Bibliographical References}

\bibliographystyle{lrec2026-natbib}
\bibliography{citations/longitudinal_clinical,citations/rad_summ_factuality,citations/cross_doc_coreference, citations/clusteringLLM, 
citations/prompts_and_agents, citations/timeline_long_text_summ, citations/embedding, citations/long_context}

\appendix
\section*{Appendix}
\section{Annotation Guidelines}
\label{section:annotation_guideline}
\begin{table*}[ht]
\centering
\renewcommand{\arraystretch}{1.3}
\begin{tabular}{|p{0.95\textwidth}|}
\hline
\textbf{Task 1: Cell-level Factuality Rating} \\ 
\hline

\begin{itemize}
    \item For all cells in the column, rate the factuality level [0 bad, 1 partially correct, 2 good] for each cell based on the consistency with the column-associated report.
    
    \item Annotators should focus only on finding-related facts written in the specific report being evaluated. Do not refer to facts from other reports.
    
    \item Correctness of those facts should \textbf{NOT} affect the rating:
    \begin{itemize}
        \item Suggestive conditions
        \begin{itemize}
            \item e.g., ``considering neoplasm given the persistent nodules''
            \item e.g., ``likely due to infection''
        \end{itemize}
        
        \item Facts that are not lung-related or not radiological findings
        \begin{itemize}
            \item e.g., ``Lung RAD rating is 3A''
            \item e.g., ``Coronary Artery Calcifications''
        \end{itemize}
    \end{itemize}
    
    \item \textbf{Skip `not-mentioned' cells.} Blank cells are equivalent to `not mentioned'. Instead of annotating them, add them in Task 2 for missing findings if actually mentioned.
    
    \item Label the cell as not containing lung-related findings as \textbf{[-1: non-lung-findings]}.
    
    \item Add a note column next to the factuality cell.
\end{itemize} \\

\hline
\end{tabular}
\caption{Annotation guidelines for factuality rating 
}
    \label{tab:annotation_step1}

\end{table*}

\begin{table*}[ht]
\centering
\renewcommand{\arraystretch}{1.3}
\begin{tabular}{|p{0.95\textwidth}|}
\hline
\textbf{Task 2: Group Assignments} \\ 
\hline

Assign a \textbf{group label} for each cell. 
The group labels to choose are the headers in the \texttt{finding\_group} column. 
You may \textbf{add a new group label} if no existing group labels can distinguish this cell from other groups. \\

What should be included in a group  
\begin{itemize}
    \item The same finding in the notes, even if described differently, \textbf{or}
    \item Findings from different notes that can change to one another.
\end{itemize}
This may lead to trend descriptions such as:
\begin{itemize}
    \item new
    \item increasing
    \item resolving
\end{itemize} \\

Duplicate labels 
If two group labels are duplicated, \textbf{choose one label and use it consistently} for all cells belonging to that group. \\

Goal: \\ 
Ensure that the same findings or time-related findings are placed in the same group. \\

Skip condition:  \\ 
\textbf{Skip cells where factuality is 0 or -1.} \\
\\
\hline
\end{tabular}
\caption{Annotation guidelines for group assignments 
}
    \label{tab:annotation_step2}
\end{table*}

\begin{table*}[ht]
\centering
\renewcommand{\arraystretch}{1.3}
\begin{tabular}{|p{0.95\textwidth}|}
\hline
\textbf{Task 3: Group Name Revision} \\ 
\hline
Given a group of dated findings, add the group name for this group. 
The goal is to make the group name clearly distinguishable from other groups while remaining concise and containing critical information that a radiologist might use when referring to the finding. \\

Strategy: \\
The group name should:
\begin{enumerate}
    \item \textbf{Include the facts shared by all members in the group.} 
    When paraphrasing occurs, choose the description used more frequently.
    
    \item \textbf{Omit details that can change in the future} 
    (e.g., size, change trend, severity).
    
    \item Use extra details if needed to distinguish similar groups.
\end{enumerate} \\

\hline
\end{tabular}
\caption{Annotation guidelines for group name revision given formed groups. }
    \label{tab:annotation_step3}

\end{table*}

\begin{figure*}[htbp]
    \centering
    \includegraphics[width=\textwidth,trim=0 200pt 0 0, clip]{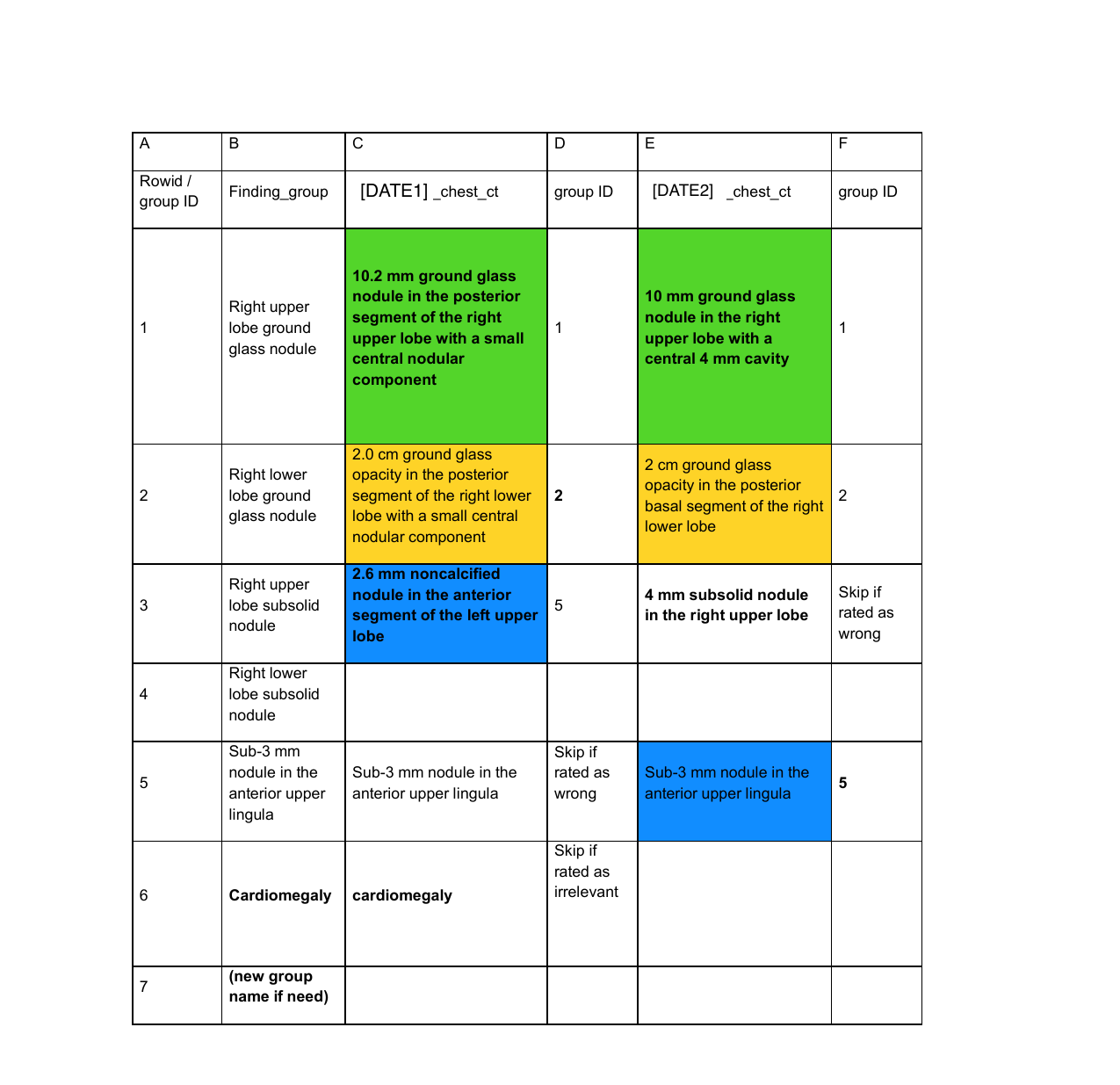}
    \caption{A group assignment example provided to annotators.}
    \label{fig:step2_example}
\end{figure*}

The annotation pipeline consists of three sequential tasks.
In the first task (Table~\ref{tab:annotation_step1}),
each cell is rated for factuality into levels of -1, 0, 1, 2.
The second task (Table~\ref{tab:annotation_step2}),
collects final task group assignment on cells to form the timeline rows. Figure~\ref{fig:step2_example} 
shows one example presented to annotators.
Finally, the third task (Table~\ref{tab:annotation_step3})
asks an annotator to revise the group names on formed groups.

\section{Prompts}
\label{section:prompt}

The prompts used in the three-step pipeline are provided in the subsequent tables. The step 1 prompt for finding extraction  is shown in Table \ref{tab:step1_extraction_prompt}, and the step 2 group name generation prompt is presented in Table \ref{tab:step2_group_name_generation_list_prompt}.

Step 3 group assignment prompts have three versions: single finding (Table~\ref{tab:step3_single_finding}), multi-finding (Table~\ref{tab:step3_multi_finding}), and multi-finding without group names provided (Table~\ref{tab:step3_group_assignment_no_groups}).

\begin{table}[t]
  \centering
  \small
  \begin{tabular}{p{0.94\linewidth}}
    \hline
    \textbf{Prompt for step 1 finding extraction} \\
    \hline

    \textbf{Q:} 
\textcolor{gray}{[\textbf{Single\_Radiology\_Report\_Text}, e.g. EXAMINATION:
CT of the chest without contrast - low dose lung cancer screening..... 
]}: \\

Please list all lung-related findings (including pleura-ralated) from the exam report.
        Please use a python list to include the strings of those lung desription. 
        Each string should include all details related to higher or lower risk, using the same description in the note , especially finding’s location, size, trend of change, and shape, density, or other characteristics. \\
        Good  example:  
          [“stable 2 mm left upper lobe calcified nodule”, 'left lower lobe opacities']\\ 
        Now please anwswer: \\
        lung\_finding\_list = \\

            \textbf{A:} Here is the list of lung-related findings from the exam report:

```python
lung\_finding\_list = [
    "2.6 mm noncalcified nodule in the anterior segment of the left upper lobe",
    "alveolar groundglass opacity in the posterior segment of the right upper lobe with a small central nodular component measuring 10.2 mm",
    ...  \\[0.6em]

    \hline
  \end{tabular}
  \caption{Step 1 finding extraction input. The prompt includes the single report input, an one-shot example, and instructions requiring a python list output format.}
  \label{tab:step1_extraction_prompt}
\end{table}


\begin{table}[t]
  \centering
  \small
  \begin{tabular}{p{0.94\linewidth}}
    \hline
    \textbf{Prompt for step 2 group name generation} \\
    \hline
    \textbf{Q:} 
\textcolor{gray}{[\textbf{Longitudinal\_Textual\_Reports}, e.g. date,note\_tag,QOClassificationName,ReportText
YYYY-MM-DD,chest\_ct,CT Lung Low Dose Screening,"EXAMINATION:
CT of the chest without contrast - low dose lung cancer screening  ... 
]}: \\

 Please group the lung-related findings recorded over time (including pleura-related findings) and list the group names. \\
Each group should contain findings that are essentially the same, even if described differently, as well as findings that evolve into one another over time. \\
Different findings or findings will not evolve into one another should be in separate groups \\

Group names should be specific that separate unrelated findings, avoid generic names. \\
       Good group name: right upper lobe ground glass nodule \\
       bad group name: ground glass nodules \\
       Finally give me a list of group names separated by commas ',' \\

         \textbf{A:} ....  List of group names: Right upper lobe ground glass nodule, Right lower lobe ground glass opacity with central nodular component, ... \\
    \hline
  \end{tabular}
  \caption{Step 2 group name generation prompt that requires list group names at the end of response. The prompt includes a good and a bad example for individual group names, but no examples for the list format.}
\label{tab:step2_group_name_generation_list_prompt}
\end{table}

\begin{table}[t]
  \centering
  \small
  \begin{tabular}{p{1\linewidth}}
    \hline
    \textbf{Prompt for step 3 group assignment for single finding } \\
    \hline
  \textbf{Q:} {\color{gray} [\underline{\textbf{Longitudinal\_Textual\_Reports}}, e.g. date,note\_tag,QOClassificationName,ReportText
YYYY-MM-DD,chest\_ct,CT Lung Low Dose Screening,"EXAMINATION:
CT of the chest without contrast - low dose lung cancer screening  ... 
] }\\

            Please find  the group best describing this finding  given the \underline{radiology exam history and} group labels.

            There are groups presented to track the lung-related findings recorded over time.   \\
            Each group should contain findings that are essentially the same, even if described differently, as well as findings that evolve into one another over time. \\
            Different findings or findings will not evolve into one another should be in separate groups \\

\vspace{6pt}
   \textcolor{gray}{ \textbf{[Zero-Shot Content]}} \\
            The group labels are: \\ 
            
           { \color{gray}[\textbf{Group name list from step 2:}  e.g.  Right upper lobe ground glass nodule, Right lower lobe ground glass nodule, ...]}, other   \\ 
            The finding to be classified is:{ \color{gray} [\textbf{Single finding from step 1:} e.g.   alveolar groundglass opacity in the posterior segment of the right upper lobe with a small central nodular component measuring 10.2 mm]}

            Give me the group in the last line starting with 'Answer:'. \\

\vspace{6pt}
   \textcolor{gray}{ \textbf{[Few-Shot Content]}} \\

 Here are some examples:

            Example 1

            Group labels: small RUL nodule, 10 mm right lower lobe nodule, other

            Finding to be classified: Small 4 mm right upper lobe nodule stable

            Answer: small RUL nodule

            Example 2

            Group labels: calcified granuloma, right upper lobe nodules, other

            Finding to be classified: Benign appearing calcified granuloma in left lung

            Answer: calcified\_granuloma

            Your task:

            Group labels:  { \color{gray}[\textbf{Group name list from step 2:}  e.g.  Right upper lobe ground glass nodule, Right lower lobe ground glass nodule, ...]}, other

            Finding to be classified: { \color{gray} [\textbf{Single finding from step 1:} e.g.   alveolar groundglass opacity in the posterior segment of the right upper lobe with a small central nodular component measuring 10.2 mm]}

            Your explain: [EXPLAIN WHY]

            Answer: [GROUP\_NAME CHOICE ]

            Always end with "Answer: [GROUP\_NAME]" as the very last line  \\
         \textbf{A:} ...  Answer: Right upper lobe ground glass nodule\\
    \hline
  \end{tabular}
  \caption{Step 3 Group Assignment for Single Finding. The prompts include underlined texts only when using full-contexts. We have both zero-shot and few-shot instructions, few-shot version adds two examples and further emphasize on the format requirements. }
  \label{tab:step3_single_finding}
\end{table}

\begin{table}[t]
  \centering
  \small
  \begin{tabular}{p{0.94\linewidth}}
    \hline
    \textbf{Prompt for step 3 group assignment (multi-finding tag format, no group names given) } \\
    \hline
  \textbf{Q:} Task:  

            Your task is to group radiology findings evolving over time from multiple reports of the same patient. \\
            Please replace every <\#> with a group name tag in angle-bracket format, such as <10\_mm\_right\_lower\_lobe\_nodule>.  \\
            You should invent your own group tags so that each tag is representative for its associated group. \\
            Use the same group name tag across all reports for all mentions related to the same underlying entity. \\
            
            Your Task: \\
            Radiology Findings: \\

{\color{gray}\textbf{[Finding\_List\_with\_Markers}}, \\
       {\color{gray} e.g. 
        2.6 mm noncalcified nodule in the anterior segment of the left upper lobe <\#>} \\
  {\color{gray}alveolar groundglass opacity in the posterior segment of the right upper lobe with a small central nodular component measuring 10.2 mm <\#> ...]} \\

            Radiology Findings END \\

            Please apply a group name tag consistently to all members of the group. \\

            Please reply with exactly the same lines listed above that are between ‘Radiology Findings’ and ‘Radiology Findings END’, but with the group name tags inserted.  \\

    \hline
  \end{tabular}
  \caption{Step 3 Group Assignment for Multi-Finding when no group names are given. The prompt includes an example of single tag, but no examples for the final output for all provided findings.  }
  \label{tab:step3_group_assignment_no_groups}
\end{table}

\begin{table*}[t]
  \centering
  \small
  \renewcommand{\arraystretch}{0.8}
  \begin{tabular}{p{1.0\linewidth}}
    \hline
    \textbf{Prompt for step 3 group assignment for multi finding } \\
    \hline
  \textbf{Q:} {\color{gray} [\underline{\textbf{Longitudinal\_Textual\_Reports}}, e.g. date,note\_tag,QOClassificationName,ReportText
YYYY-MM-DD,chest\_ct,CT Lung Low Dose Screening,"EXAMINATION:
CT of the chest without contrast - low dose lung cancer screening  ... 
] }\\
            Task: 
            Your task is to group radiology findings evolving over time from multiple reports of the same patient. \underline{Those reports are as given above.} \\
            Please replace every <\#> with a group name tag in angle-bracket format, such as <10\_mm\_right\_lower\_lobe\_nodule>. \\
            You can use either an existing group tag as provided, or invent your own. \\
            Use the same group name tag across all reports for all mentions related to the same underlying entity. \\

\vspace{6pt}
   \textcolor{gray}{ \textbf{[Few-Shot Content]}} \\
            Here are examples:\\
            -------------------------- \\
            Example 1 \\
            Existing group name tags:  \\
            <10\_mm\_right\_lower\_lobe\_nodule> : 10 mm right lower lobe nodule \\
            <small\_RUL\_nodule> : small RUL nodule \\
            Radiology Findings: \\
            Small 4 mm right upper lobe nodule is new <\#>  \\
            The 10 mm right lower lobe groundglass is stable <\#> \\
            Here are several left lobe calcified nodules <\#> \\
            Small right upper lobe nodule is stable <\#> \\
            RLL groundglass is smaller now <\#> \\
            Radiology Findings END \\
            Answer:  \\
            Small 4 mm right upper lobe nodule is new <small\_RUL\_nodule> \\
            The 10 mm right lower lobe groundglass is stable <10\_mm\_right\_lower\_lobe\_nodule> \\
            Here are several left lobe calcified nodules <left\_lobe\_calcified\_nodules> \\
            Small right upper lobe nodule is stable <small\_RUL\_nodule> \\
            RLL groundglass is smaller now <10\_mm\_right\_lower\_lobe\_nodule> \\
            -------------------------- \\
            Example 2 \\
            Existing group name tags: \\
            <calcified\_granuloma> : calcified granuloma \\
            <right\_upper\_lobe\_nodules> : right upper lobe nodules  \\
            Radiology Findings: \\
            Benign appearing calcified granuloma in the left lung <\#> \\
            New right upper lobe nodules compared to last scan <\#> \\
            A 1.5 cm part-solid nodule is identified in the lingula <\#> \\
            Radiology Findings END \\
            Answer: \\
            Benign appearing calcified granuloma in the left lung  <calcified\_granuloma> \\
            New right upper lobe nodules compared to last scan <right\_upper\_lobe\_nodules> \\
            A 1.5 cm part-solid nodule is identified in the lingula <part\_solid\_nodule\_in\_the\_lingula> \\
            -------------------------- \\
          
   \textcolor{gray}{ \textbf{[Few-Shot Content END]}} \\
   \vspace{6pt}

            Your Task: \\

            Existing group name tags: \\
             
           { \color{gray}[\textbf{Group name list from step 2:}  e.g.  right\_upper\_lobe\_ground\_glass\_nodule, right\_lower\_lobe\_ground\_glass\_nodule, ...]} \\

            Radiology Findings: \\
            
[{\color{gray}\textbf{Finding\_List\_with\_Tags}}, \\
       {\color{gray} e.g. 
        2.6 mm noncalcified nodule in the anterior segment of the left upper lobe <\#>} \\
  {\color{gray}alveolar groundglass opacity in the posterior segment of the right upper lobe with a small central nodular component measuring 10.2 mm <\#> ...]} \\

            Radiology Findings END \\

            If an existing group name tag fits, use it; \\
            If multiple existing tags fit, choose one and apply it consistently to all members of the group. \\
            If no existing group name fits, invent your own group tag. \\
            Please reply with exactly the same lines listed above that are between ‘Radiology Findings’ and ‘Radiology Findings END’, but with the group name tags inserted.            \\
            Answer: \\
    \hline
  \end{tabular}
  \caption{Step 3 Group Assignment for Multi-Finding when group names are given. The prompts include underlined texts only when using full-contexts. We have both zero-shot and few-shot instructions, few-shot version adds two examples. }
  \label{tab:step3_multi_finding}
\end{table*}
  \renewcommand{\arraystretch}{1}

\section{Qualitative Analysis}
\label{section:examples}
We present LLM generation examples for Step 1 finding extraction in Table~\ref{tab:step1_examples} and Step 2 group name generation in Table~\ref{tab:step2_examples} and Table~\ref{tab:step2_examples_full_vs_list}. In both steps, GPT-4o generates fewer findings with greater conciseness, while Llama 3.1 tends to generate more findings and include content irrelevant to lung radiological findings.
In Step 2, when switching from full longitudinal report context to the predicted finding list as a shorter context, GPT-4o produces more group names, possibly because the full context helps merge findings. However, this trend of reduced group names is not observed in Llama 3.1, potentially because the shorter context instead reduces irrelevant content in the Llama output.
\begin{table*}[htbp]
\centering
\small
\begin{tabular}{|p{5cm}|p{5cm}|p{5cm}|}
\hline
\textbf{Gold} & \textbf{GPT-4o} & \textbf{Llama3.1} \\
\hline
Moderate centrilobular emphysema & Moderate centrilobular emphysema & Moderate centrilobular emphysema \\

stable lung nodules since [DATE], therefore benign. No new or enlarging pulmonary nodule & Stable lung nodules since [DATE], therefore benign & stable lung nodules since [DATE], therefore benign. No new or enlarging pulmonary nodule \\

 & No new or enlarging pulmonary nodule & Normal supraclavicular region \\

 &  & Normal mediastinum \\

 &  & Normal pleura \\

 &  & Mildly dilated main pulmonary artery measuring 3.3 cm in caliber, also unchanged \\

 &  & Stable mildly dilated pulmonary artery may represent pulmonary hypertension \\

 &  & Stable mildly dilated ascending aorta may represent systemic hypertension \\
\hline
Lungs: A few tiny (1--2 mm) nodules are scattered throughout the lungs [IMAGE NUMBES]. & No large masses or dense airspace consolidations & Supraclavicular region: Normal. \\

 & A few tiny (1--2 mm) nodules are scattered throughout the lungs & Mediastinum: Normal. \\

 &  & Lungs: No large masses or dense airspace consolidations. \\

 &  & Lungs: A few tiny (1--2 mm) nodules are scattered throughout the lungs. \\

 &  & Pleura: Normal. \\
\hline

Low lung volume & low lung volume with left base atelectasis & Cardiomegaly \\

Left base atelectasis & no evidence of edema & Low lung volume \\

 & no pneumothorax & Left base atelectasis \\

 &  & No evidence of edema \\

 &  & No pneumothorax \\
\hline
Right basilar consolidation likely pneumonia & patchy consolidation at the left lung base & Patchy consolidation at the left lung base \\

Patchy consolidation at the left lung base & minimal density at right lung base & Minimal density at right lung base \\

Left basilar consolidation likely pneumonia &  & Right basilar consolidation likely pneumonia \\

Minimal density at right lung base &  & Left basilar consolidation likely pneumonia \\
\hline

predominant fibrosis as before & predominant fibrosis & predominant fibrosis \\

increased opacification of the left lower lobe & increased opacification of the left lower lobe best seen in the lateral view which may represent atelectasis, aspiration, or pneumonia & increased opacification of the left lower lobe \\

 &  & atelectasis, aspiration, or pneumonia (left lower lobe) \\

 &  & no pleural effusion \\

 &  & no pneumothorax \\
\hline
\end{tabular}
\caption{Step 1 finding extraction examples from GPT-4o and Llama 3.1 8B Instruct. Llama 3.1 tends to generate more findings, and to include content irrelevant to lung findings. Human references are provided as the gold labels.
The predicted findings are  presented in the same order as they appear in the LLM responses.}
\label{tab:step1_examples}
\end{table*}

\begin{table*}[htbp]
\centering
\small
\begin{tabular}{|p{4cm}|p{4cm}|p{4cm}|p{4cm}|}
\hline
\textbf{Gold  (group assignments from  Ann1)} & \textbf{Gold  (group assignments from  Ann2)} & \textbf{GPT-4o} & \textbf{Llama3.1} \\
\hline

Hyperinflated lungs 
& Centrilobular emphysema in upper lobe predominant distribution 
& Left upper lobe spiculated nodule 
& Centrilobular emphysema in upper lobe predominant distribution \\

Flattening of domes of diaphragm 
& Focal area of scarring 
& Centrilobular emphysema 
& Spiculated nodule in left upper lobe (concerning for primary lung malignancy) \\

Centrilobular emphysema in upper lobe predominant distribution 
& Spiculated nodule in left upper lobe (concerning for primary lung malignancy) 
& Pleura and pleural spaces 
& Unchanged spiculated left upper lobe nodule (6 x 9 mm) \\

Left anterior lingula scarring 
& Right upper lobe nodule 
& Pericardial effusion 
& Unchanged right upper lobe nodule (2 mm) \\

Spiculated nodule in left upper lobe (concerning for primary lung malignancy) 
&  
& Ascending aorta abnormalities 
& No new pulmonary nodules \\

Right upper lobe nodule 
&  
&  
& Pleural effusions or pneumothorax (no findings) \\

Bilateral upper lobe lung nodules 
&  
&  
& Small bilateral fat-containing Bochdalek hernias \\

&  
&  
& Mild enlargement of the left lobe of the thyroid (supraclavicular region) \\

&  
&  
& Normal heart size \\

&  
&  
& Trace pericardial effusion (likely physiologic) \\

&  
&  
& Ascending aorta ectasia (unchanged) \\

&  
&  
& Mildly dilated main pulmonary artery (stable) \\

&  
&  
& Nonobstructive bilateral punctate renal calculi (upper abdomen) \\

&  
&  
& Moderate centrilobular emphysema \\

&  
&  
& Large lung volumes consistent with emphysema \\

&  
&  
& No acute bone or soft tissue abnormality \\

\hline
\end{tabular}
\caption{Step 2 group name generation examples  from GPT-4o and Llama 3.1 8B Instruct when using full longitudinal reports as the context. Gold group names are provided in two versions, each reflecting the grouping assignments of Ann1 and Ann2 respectively, with names refined by a single third annotator across both versions. Similarly to step 1 results, Llama 3.1 tends to generate more group names and to include content irrelevant to lung findings. Two versions of human references are provided as the gold labels. The predicted group names are  presented in the same order as they appear in the LLM responses.}
\label{tab:step2_examples}
\end{table*}

\begin{table*}[htbp]
\centering
\small
\begin{tabular}{|p{4cm}|p{4cm}|p{4cm}|p{4cm}|}
\hline
\textbf{GPT-4o (Full)} & \textbf{GPT-4o (List)} & \textbf{Llama3.1 (Full)} & \textbf{Llama3.1 (List)} \\
\hline

Left upper lobe spiculated nodule 
& Hyperinflated lungs 
& Centrilobular emphysema in upper lobe predominant distribution 
& Hyperinflated lungs and diaphragm flattening \\

Centrilobular emphysema 
& Flattening of the domes of the diaphragm 
& Spiculated nodule in left upper lobe (concerning for primary lung malignancy) 
& Centrilobular emphysema \\

Pleura and pleural spaces 
& Centrilobular emphysema 
& Unchanged spiculated left upper lobe nodule (6 x 9 mm) 
& Spiculated nodule in left upper lobe (primary lung malignancy concern) \\

Pericardial effusion 
& Spiculated nodule in the left upper lobe 
& Unchanged right upper lobe nodule (2 mm) 
& Small focal area of scarring in lingula \\

Ascending aorta abnormalities 
& Right upper lobe pulmonary nodule 
& No new pulmonary nodules 
& Clear trachea and central airways \\

& Small focal area of scarring within the anterior aspect of the lingula 
& Pleural effusions or pneumothorax (no findings) 
& Pleural effusions or pneumothorax (absence of) \\

& No pleural effusions or pneumothorax 
& Small bilateral fat-containing Bochdalek hernias 
& Bilateral fat-containing Bochdalek hernias \\

& Small bilateral fat-containing Bochdalek hernias 
& Mild enlargement of the left lobe of the thyroid (supraclavicular region) 
& Large lung volumes consistent with emphysema \\

& Large lung volumes consistent with emphysema 
& Normal heart size 
& Stable lung nodules (benign) \\

& The lungs and pleural spaces are clear 
& Trace pericardial effusion (likely physiologic) 
& Normal supraclavicular region and mediastinum \\

&  
& Ascending aorta ectasia (unchanged) 
& Normal pleura \\

&  
& Mildly dilated main pulmonary artery (stable) 
& Mildly dilated main pulmonary artery (pulmonary hypertension concern) \\

&  
& Nonobstructive bilateral punctate renal calculi (upper abdomen) 
& Mildly dilated ascending aorta (systemic hypertension concern) \\

&  
& Moderate centrilobular emphysema 
&  \\

&  
& Large lung volumes consistent with emphysema 
&  \\

&  
& No acute bone or soft tissue abnormality 
&  \\

\hline
\end{tabular}
\caption{Comparison of step 2 group name generation examples using full longitudinal reports versus predicted finding lists as context. Examples are from GPT-4o and Llama 3.1 8B Instruct. The predicted group names are  presented in the same order as they appear in the LLM responses.}
\label{tab:step2_examples_full_vs_list}
\end{table*}

\end{document}